\documentclass{article}

\usepackage[final,nonatbib]{neurips_2019}

\usepackage[utf8]{inputenc} 
\usepackage[T1]{fontenc}    
\usepackage{hyperref}       
\usepackage{url}            
\usepackage{booktabs}       
\usepackage{amsfonts}       
\usepackage{nicefrac}       
\usepackage{microtype}      
\usepackage{amsmath,amsfonts}
\usepackage{graphicx}
\usepackage{sidecap}
\usepackage{subfig}
\usepackage{xcolor}
\usepackage{placeins}

\title{Inherent Weight Normalization in Stochastic Neural Networks}

\author{%
   Georgios Detorakis \\
   Department of Cognitive Sciences \\
   University of California Irvine \\
   Irvine, CA 92697 \\
   \texttt{gdetorak@uci.edu} 
   \And
   Sourav Dutta \\
   Department of Electrical Engineering \\
   University of Notre Dame \\
   Notre Dame, IN 46556 USA \\
   \texttt{sdutta4@nd.edu} \\
   \AND
   Abhishek Khanna \\
   Department of Electrical Engineering \\
   University of Notre Dame \\
   Notre Dame, IN 46556 USA    \\
   \texttt{akhanna@nd.edu} 
   \And
   Matthew Jerry \\
   Department of  Electrical Engineering \\
   University of Notre Dame \\
   Notre Dame, IN 46556 USA    \\
   \texttt{mjerry@alumni.nd.edu} \\
   \AND
   Suman Datta \\
   Department of Electrical Engineering  \\
   University of Notre Dame \\
   Notre Dame, IN 46556 USA    \\
   \texttt{sdatta@nd.edu} 
  \And
   Emre Neftci \\
   Department of Cognitive Sciences\\
   Department of Computer Science\\
   University of California Irvine\\
   Irvine, CA 92697 \\
   \texttt{eneftci@uci.edu} \\
}

\DeclareMathOperator{\sgn}{sgn}

\begin{document}

\maketitle

\begin{abstract}
Multiplicative stochasticity such as Dropout improves the robustness and generalizability of deep neural networks.
Here, we further demonstrate that always-on multiplicative stochasticity combined with simple threshold neurons are sufficient operations for deep neural networks.
We call such models Neural Sampling Machines (NSM). 
We find that the probability of activation of the NSM exhibits a self-normalizing property that mirrors Weight Normalization, a previously studied mechanism that fulfills many of the features of Batch Normalization in an online fashion.
The normalization of activities during training speeds up convergence by preventing internal covariate shift caused by changes in the input distribution.
The always-on stochasticity of the NSM confers the following advantages: the network is identical in the inference and learning phases, making the NSM suitable for {online learning}, it can exploit stochasticity inherent to a physical substrate such as analog non-volatile memories for in-memory computing, and it is suitable for Monte Carlo sampling, while requiring almost exclusively addition and comparison operations.
We demonstrate NSMs on standard classification benchmarks (MNIST and CIFAR) and event-based classification benchmarks (N-MNIST and DVS Gestures). Our results show that NSMs perform comparably or better than conventional artificial neural networks with the same architecture.
\end{abstract}

\section{Introduction}

Stochasticity is a valuable resource for computations in biological and 
artificial neural networks~\cite{Buesing_etal11,Moreno-Bote14,Al-Shedivat_etal15a}. 
It affects neural networks in many different ways. Some of them are
(i) escaping local minima during learning and inference~\cite{Ackley_etal85}, (ii)
stochastic regularization in neural networks~\cite{Hinton_etal12,Wan_etal13}, (iii)
Bayesian inference approximation with Monte Carlo sampling~\cite{Buesing_etal11,Gal_Ghahramani15}, 
(iv) stochastic facilitation~\cite{McDonnell_Ward11}, and (v) energy efficiency in
computation and communication~\cite{Levy_Baxter02,Harris_etal12}. 

In artificial neural networks, multiplicative noise is applied as random variables that multiply network weights or neural activities (\emph{e.g.} Dropout).
In the brain, multiplicative noise is apparent in the probabilistic nature of neural activations \cite{Harris_etal12} and their synaptic quantal release~\cite{branco:2009,walmsley:1987}. 
Analog non-volatile memories for in-memory computing such as resistive RAMs, ferroelectric devices or phase-change materials~\cite{Yu_etal16,Jerry_etal17c,Eryilmaz_etal16} exhibit a wide variety of stochastic behaviors \cite{Querlioz_etal15,Naous_etal16,Yu_etal16,Mulaosmanovic_etal17}, including set/reset variability \cite{Ambrogio_etal14} and random telegraphic noise \cite{Ambrogio_etal14b}.
In crossbar arrays of non-volatile memory devices designed for vector-matrix multiplication (\emph{e.g.} where weights are stored in the resistive or ferroelectric states), such stochasticity manifests itself as multiplicative noise.

Motivated by the ubiquity of multiplicative noise in the physics of artificial and biological computing substrates, we explore here Neural Sampling Machines (NSMs): a class of neural networks with binary threshold neurons that rely almost exclusively on multiplicative noise as a resource for inference and learning. 
We highlight a striking self-normalizing effect in the NSM that fulfills a role that is similar to Weight Normalization during learning \cite{Salimans_Kingma16}.
This normalizing effect prevents internal covariate shift as with Batch Normalization \cite{Ioffe_Szegedy15}, stabilizes the weights distributions during learning, and confers rejection to common mode fluctuations in the weights of each neuron.



We demonstrate the NSM on a wide variety of classification tasks, including classical benchmarks and neuromorphic, event-based benchmarks.
The simplicity of the NSM and its distinct advantages make it an attractive model for hardware implementations using non-volatile memory devices. 
While stochasticity there is typically viewed as a disadvantage, the NSM has the potential to exploit it.
In this case, the forward pass in the NSM simply boils down to weight memory lookups, additions, and comparisons.

\subsection{Related Work}

The NSM is a stochastic neural network with discrete binary units and thus closely related to Binary Neural Networks (BNN).
BNNs have the objective of reducing the computational and memory footprint of deep neural networks at run-time~\cite{Courbariaux_etal16,Rastegari_etal16}.
This is achieved by using binary weights and simple activation functions that require only bit-wise operations. 

Contrary to BNNs, the NSM is stochastic during both inference and learning.  
Stochastic neural networks are argued to be useful in learning multi-modal distributions and conditional computations~\cite{Bengio_etal13,Tang_Salakhutdinov13} and for encoding uncertainty~\cite{Gal_Ghahramani15}.

Dropout and Dropconnect techniques randomly mask a subset of the neurons and the connections during train-time for regularization and preventing feature co-adaptation~\cite{Hinton_etal12,Wan_etal13}. 
These techniques continue to be used for training modern deep networks.
Dropout during inference time can be viewed as approximate Bayesian inference in
deep Gaussian processes~\cite{Gal_Ghahramani15}, and this technique is referred
to as Monte Carlo (MC) Dropout.
NSMs are closely related to MC Dropout, with the exception that the activation function is stochastic and the neurons are binary.
Similarly to MC Dropout, the ``always-on'' stochasticity of NSMs can be in principle articulated as a MC integration over an equivalent Gaussian process posterior approximation, fitting the predictive mean and variance of the data. 
MC Dropout can be used for active learning in deep neural networks, whereby a
learner selects or influences the training dataset in a way that optimally 
minimizes a learning criterion~\cite{Gal_Ghahramani15,Cohn_etal95}.

Taken together, NSM can be viewed as a combination of stochastic neural networks, Dropout and BNNs.
While stochastic activations in the binarization function are argued to be inefficient due to the generation of random bits, stochasticity in the NSM, however, requires only one random bit per pass per neuron or per connection. 
A different approach for achieving zero mean and unit variance is the self-normalizing neural networks proposed
in~\cite{klambauer:2017}. 
There, an activation function in non-binary, deterministic networks is constructed mathematically so that outputs are normalized.
In contrast, in the NSM unit, normalization in the sense of \cite{Salimans_Kingma16} emerges from the multiplicative noise as a by-product of the central limit theorem.
This establishes a connection between exploiting the physics of hardware systems and recent deep learning techniques, while achieving good accuracy on benchmark classification tasks.
Such a connection is highly significant for the devices community, as it implies a simple circuit (threshold
operations and crossbars) that can exploit (rather than mitigate) device non-idealities such as read stochasticity.


In recurrent neural networks, stochastic synapses were shown to behave as stochastic counterparts of Hopfield networks~\cite{Neftci_etal16}, but where stochasticity is caused by multiplicative noise at the synapses (rather than logistic noise in Boltzmann machines).
These were shown to surpass the performances of equivalent machine learning algorithms~\cite{Hinton02,Neftci_etal14} on certain benchmark tasks. 

\subsection{Our Contribution}

In this article, we demonstrate multi-layer and convolutional neural networks employing NSM layers on GPU simulations, and compare with their equivalent deterministic neural networks. We articulate NSM's self-normalizing effect as a statistical equivalent of Weight Normalization.
Our results indicate that a neuron model equipped with a hard-firing threshold
(\emph{i.e.}, a Perceptron) and stochastic neurons and synapses:
\begin{itemize}
    \item Is a sufficient resource for stochastic, binary deep neural networks.
    \item Naturally performs weight normalization.
    \item Can outperform standard artificial neural networks of comparable size.
\end{itemize}

The always-on stochasticity gives the NSM distinct advantages compared to traditional deep neural networks or binary neural networks:
The shared forward passes for training and inference in NSM are consistent with the requirement of {online learning since an NSM implements weight normalization, which is not based on batches~\cite{Salimans_Kingma16}}.
This enables simple implementations of neural networks with emerging devices.
Additionally, we show that the NSM provides robustness to fluctuations and fixed precision of the weights during learning.

\subsection{Applications}
During inference, the binary nature of the NSM equipped with blank-out noise makes it largely multiplication-free. As with the Binarynet~\cite{Courbariaux_Bengio16} or XNORnet~\cite{Rastegari_etal16}, we speculate that they can be most advantageous in terms of energy efficiency on dedicated hardware.

The NSM is of interest for hardware implementations in memristive crossbar arrays, as threshold units are straightforward to implement in CMOS and binary inputs mitigate read and write non-idealities in emerging non-volatile memory devices while reducing communication bandwidth \cite{Yu_etal16}.
Furthermore, multiplicative stochasticity in the NSM is consistent with the stochastic properties of emerging nanodevices~\cite{Querlioz_etal15,Naous_etal16}. 
Exploiting the physics of nanodevices for generating stochasticity can lead to significant improvements in embedded, dedicated deep learning machines.

\section{Methods}

\subsection{Neural Sampling Machines (NSM)}
We formulate the NSM as a stochastic neural network model that exploits the properties of multiplicative noise to perform inference and learning. 
For mathematical tractability, we focus on threshold (sign) units, where 
$\mathrm{sgn}: \mathbb{R} \rightarrow [-1,1]$, 
\begin{align}
    \label{eq:hopfield-activation}
  z_i = \mathrm{sgn}(u_i) = \begin{cases} 1 &\mbox{if } u_i \ge 0 \\
        -1 & \mbox{if } u_i < 0 \end{cases},
\end{align}
where $u_i$ is the pre-activation of neuron $i$ given by the following equation
\begin{align}
    \label{eq:preactiv}
  u_i = \sum_{j=1}^N \xi_{ij}  w_{ij} z_j + b_i + \eta_{i},
\end{align}
where $\xi_{ij}$ and $\eta_{i}$ represent multiplicative and additive noise
terms, respectively. Both $\xi$ and $\eta$ are independent and identically 
distributed (\emph{iid}) random variables. $w_{ij}$ is the weight of the connection between 
neurons $i$ and $j$, $b_i$ is a bias term, and $N$ is the number
of input connections (fan-in) to neuron $i$.
Note that multiplicative noise can be introduced at the synapse ($\xi_{ij}$), or at the neuron ($\xi_{i}$).

Since the neuron is a threshold unit, it follows that 
$P(z_i  = 1| \mathbf{z} ) = P(u_i  \ge 0| \mathbf{z})$.
Thus, the probability that unit $i$ is active given the network state is equal
to one minus the cumulative distribution function of $u_i$.
Assuming independent random variables $u_i$, the central limit theorem indicates
that the probability of the neuron firing is $P(z_i=1|{\bf z}) = 1 - \Phi(u_i|{\bf
z})$ (where $\Phi$ is the cumulative distribution function of normal
distribution) and more precisely
\begin{align}
  \label{eq:general_activation_function}
     P(z_i = 1| \mathbf{z}) = \frac12 \left( 1 + \operatorname{erf}
     \left( \frac{\mathbb{E}(u_i|\mathbf{z})}{\sqrt{2 \mathrm{Var}(u_i|\mathbf{z})}}\right) \right),
\end{align}
where $\mathbb{E}(u_i)$ and $\mathrm{Var}(u_i)$ are the expectation and variance
of state $u_i$.

\setlength{\abovedisplayskip}{3pt}
\setlength{\belowdisplayskip}{3pt}
\setlength{\abovedisplayshortskip}{3pt}
\setlength{\belowdisplayshortskip}{3pt}

In the case where only independent additive noise is present, equation~\eqref{eq:preactiv} is
rewritten as $ u_i = \sum_{j=1}^N w_{ij} z_j + b_i + \eta_{i}$ and the 
expectation and variance are given by $\mathbb{E}(u_i|\mathbf{z}) = \sum_{j=1}^N w_{ij} z_j + b_i + \mathbb{E}(\eta)$ and $\mathrm{Var}(u_i|\mathbf{z}) = \mathrm{Var}(\eta)$, respectively. 
In this case, equation~\eqref{eq:general_activation_function} is a sigmoidal neuron 
with an $\mathrm{erf}$ activation function with constant bias 
$\mathbb{E}(\eta)$ and constant slope $\mathrm{Var}(\eta)$.
Thus, besides the sigmoidal activation function, the additive noise case does not endow the network with any extra properties.

In the case of multiplicative noise, equation~\eqref{eq:preactiv} becomes $u_i = \sum_{j=1}^N \xi_{ij} w_{ij} z_j + b_i$ and its expectation
and variance are given by $ \mathbb{E}(u_i|\mathbf{z}) = 
\mathbb{E}(\xi)\sum_{j=1}^N w_{ij} z_j$ and 
$\mathrm{Var}(u_i|\mathbf{z}) = \mathrm{Var}(\xi) \sum_{j=1}^N w^2_{ij} $,
respectively. In this derivation, we have used the fact that the square of a 
sign function is a constant function ($\sgn^2(x) = 1$).
In contrast to the additive noise case, $\mathrm{Var}(u_i|\mathbf{z})$ is  proportional to the square of the input weight parameters. The probability of neurons being active
becomes:
\begin{align}
    \label{eq:nsm_activity_function}
    \begin{split}
    P(z_i = 1| \mathbf{z}) &= \frac12 \left(1 + \mathrm{erf}\left(  \mathbf{v}_i \cdot \mathbf{z} \right)\right), \\
      \text{with } \mathbf{v}_i &= \beta_i
      \frac{\mathbf{w}_{i}}{||\mathbf{w}_{i}||_2},
    \end{split}
\end{align}
where $\beta_i$ {here is a variable that captures the parameters of the noise process $\xi_i$}.
In the denominator, we have used the identity $\sqrt{\sum_j w_{ij}^2} = \sqrt{\sum_j w_{ij}^2 z_j^2} = ||\mathbf{w}_i||_2$, where $||\cdot||_2$ denotes the $L2$ norm of the weights of neuron $i$. 
This term has a normalizing effect on the activation function, similar to weight normalization, as discussed below.
Note that the  self-normalizing effect is not specific to the distribution of the chosen random variables, and holds as long as the random variables are \emph{iid}.

One consequence of multiplicative noise here is that any positive scaling factor applied to $\mathbf{w}_i$ is canceled out by the norm.
To counter this problem and control $\beta_i$ without changing the distribution governing $\xi$, the NSM introduces a factor $a_i$ in the preactivation's equation:
\begin{align}
    \label{eq:nsm}
     u_i = \sum_{j=1}^N (\xi_{ij} + a_i) w_{ij} z_j + b_i.
\end{align}

Thanks to the binary nature of $z_i$, equation \eqref{eq:nsm} is multiplication-free except for the term involving $a_i$. 
Since $a_i$ is defined per neuron, the multiplication operation is only performed once per neuron and time step.
In this article, we focus on two relevant cases of noise: Gaussian noise with mean $1$ and variance $\sigma^2$, $\xi_{ij}\sim \mathcal{N}(1,\sigma^2)$ and Bernoulli 
(blank-out) noise $\xi_{ij}\sim Bernoulli(p)$, with parameter $p$.
From now on we focus only on the multiplicative noise case.

\paragraph{Gaussian Noise}
In the case of multiplicative Gaussian Noise, $\xi$ in equation~\eqref{eq:nsm} is a 
Gaussian random variable $\xi\sim \mathcal{N}(1,\sigma^2)$.
This means that the expectation and variance are 
$\mathbb{E}(u_i|\mathbf{z}) = (1+a_i) \sum_j w_{ij} z_j$ and 
$\mathrm{Var}(u_i|\mathbf{z}) = \sigma^2\sum_j w_{ij}^2$, respectively. And
hence,
$
      \beta_i =  \frac{1+a_i}{\sqrt{2 \sigma^2}}.
$

\begin{figure}
    \centering
    \includegraphics[width=0.25\textwidth]{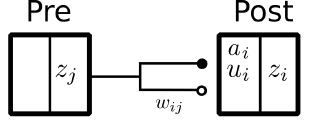}
    \caption{{\bf Blank-out synapse with scaling factors.} {Weights} are
    accumulated on $u_i$ as a sum of a deterministic term scaled by $\alpha_i$
    (filled discs) and a stochastic term with fixed blank-out probability
    $p$ (empty discs).}
    \label{fig:heterosynapse}
\end{figure}

\paragraph{Bernoulli (Blank-out) Noise}
Bernoulli (``Blank-out'') noise can be interpreted as a Dropout mask on the
neurons or a Dropconnect mask on the synaptic weights (see Fig~\ref{fig:heterosynapse}), where $\xi_{ij} \in [ 0,1]$ in equation~\eqref{eq:nsm} becomes a Bernoulli random variable with parameter $p$.
Since the $\xi_{ij}$ are independent, for a given $\mathbf{z}$, a sufficiently large fan-in, and $0<p<1$, the sums in equation~\eqref{eq:nsm} are 
Gaussian-distributed with means and variances 
$\mathbb{E}(u_i|\mathbf{z}) = (p+a_i) \sum_j w_{ij} z_j$ and
$\mathrm{Var}(u_i|\mathbf{z}) = p(1-p)\sum_j w_{ij}^2$, respectively. 
Therefore we obtain:
$
      \beta_i =  \frac{ p + a_i}{\sqrt{2 p (1-p)} }.
$

We observed empirically that whether the neuron is stochastic or the synapse is stochastic did not significantly affect the results. 

\subsection{NSMs implements Weight Normalization}
The key idea in weight normalization~\cite{Salimans_Kingma16} is to normalize unit activity by reparameterizing the weight vectors. 
The reparameterization used there has the form:
$\mathbf{v}_i = \beta_i   \frac{\mathbf{w}_{i}}{||\mathbf{w}_{i}||}$.
This is exactly the form obtained by introducing multiplicative noise in 
neurons (equation~\eqref{eq:nsm_activity_function}), suggesting that NSMs inherently 
perform weight normalization in the sense of~\cite{Salimans_Kingma16}.
The authors argue that decoupling the magnitude and the direction of the weight vectors speeds up convergence and confers many of the features of batch normalization.
To achieve weight normalization effectively, gradient descent is performed with
respect to the scalars $\mathbf{\beta}$ (which are themselves parameterized with $a_i$) in addition to the weights $\mathbf{w}$:
\begin{align}
    \label{eq:derivative_L1}
    \partial_{\beta_i} \mathcal{L} &= \frac{\sum_j w_{ij} \partial_{v_{ij}} \mathcal{L} }{||\mathbf{w}_i||}\\
    \label{eq:derivative_L2}
    \partial_{w_{ij}} \mathcal{L} &= \frac{\beta_i}{||\mathbf{w}_i||} \partial_{v_{ij}} \mathcal{L} - \frac{{\bf w}_i \beta_i}{||\mathbf{w}_i||^2} \partial_{\beta_i} \mathcal{L}  
\end{align}

\subsection{NSM Training Procedure}
Neural sampling machines (and stochastic neural networks in general) are 
challenging to train because errors cannot be directly back-propagated through
stochastic nodes. 
This difficulty is compounded by the fact that the neuron state is a discrete
random variable, and as such the standard reparametrization trick is not directly applicable~\cite{Goodfellow_etal16}.
Under these circumstances, unbiased estimators resort to minimizing expected costs through the
family of REINFORCE algorithms \cite{Williams92,Bengio_etal13,Neal90} (also called 
{score function} estimator and {likelihood-ratio} estimator).
Such algorithms have general applicability but gradient estimators have 
impractically high variance and require multiple passes in the network to 
estimate them~\cite{Raiko_etal14}.
Straight-through estimators ignore the non-linearity altogether~\cite{Bengio_etal13}, but result in networks with low performance.
Several work have introduced methods to overcome this issue, such as in discrete variational autoencoders~\cite{Rolfe16}, bias reduction techniques for the REINFORCE algorithm~\cite{Gu_etal15} and concrete distribution approach (smooth relaxations of discrete random variables)~\cite{Maddison_etal16} or other reparameterization tricks~\cite{Shayer_etal17}.

%
Striving for simplicity, here we propagate gradients through the neurons' activation probability function.
This approach theoretically comes at a cost in accuracy because the rule is a biased estimate of the gradient of the loss.
This is because the gradients are estimated using activation probability. 
However, it is more efficient than REINFORCE algorithms as it uses the information provided by the gradient back-propagation algorithm. 
In practice, we find that, provided adequate initialization, the gradients are well behaved and yield good performance while being able to leverage existing automatic differentiation capabilities of {software libraries (\emph{e.g.} gradients in Pytorch~\cite{Paszke_etal17}). In the implementation of NSMs, probabilities are only computed for the gradients in the backward pass, while only binary states are propagated in the forward pass (see SI~\ref{app:graph})}.

To assess the impact of this bias, we compare the above training method with Concrete Relaxation which is unbiased~\cite{Maddison_etal16}. 
The NSM network is compatible with the binary case of Concrete relaxation. 
We trained the NSM using BinConcrete units on MNIST data set (Test Error Rate: $0.78\%$), and observed that the angles between the gradients of the proposed NSM and BinConcrete are close (see SI~\ref{sec:concrete}). 

Unless otherwise stated, and similarly to~\cite{Salimans_Kingma16}, we use a data-dependent initialization 
of the magnitude parameters $\beta$ and the bias parameters over one batch 
of $100$ training samples such that the preactivations to each layer have zero
mean and unit variance over that batch:
\begin{align}
    \beta &\leftarrow \frac{1}{\sigma}, &
    b \leftarrow -\frac{\mu ||w|| \sqrt{2 Var(\xi)}}{\sigma},
\end{align}
where $\mu$ and $\sigma$ are feature-wise means and standard deviations 
estimated over the minibatch.
For all classification experiments, we used cross-entropy loss 
$\mathcal{L}^n = - \sum_i t_i^n \log p_{i}^n$,
where $n$ indexes the data sample and $p_{i}$ is the Softmax output.
All simulations were performed using Pytorch~\cite{Paszke_etal17}.
All NSM layers were built as custom Pytorch layers (for more details about 
simulations see SI~\ref{app:sim_details}).\footnote{\url{https://github.com/nmi-lab/neural_sampling_machines}}

\section{Experiments}

\subsection{Multi-layer NSM Outperforms Standard Stochastic Neural Networks in Speed and Accuracy}
In order to characterize the classification abilities of the NSM we trained a fully connected network on the MNIST
handwritten digit image database for digit classification. 
The network consisted of three fully-connected layers of size $300$, and a 
Softmax layer for $10$-way classification and all Bernoulli process parameters
were set to $p=.5$.
The NSM was trained using {back-propagation and a softmax layer with cross-entropy loss and minibatches of size $100$.}
As a baseline for comparison, we used the stochastic neural network ({StNN}) 
presented in~\cite{Lee_etal14} without biases, with a sigmoid activation 
probability $P_{sig}(z_i = 1| \mathbf{z}) = \mathrm{sigmoid}(\mathbf{w}_i \cdot \mathbf{z})$.

\begin{figure}
    \centering
    \includegraphics[width=0.9\textwidth]{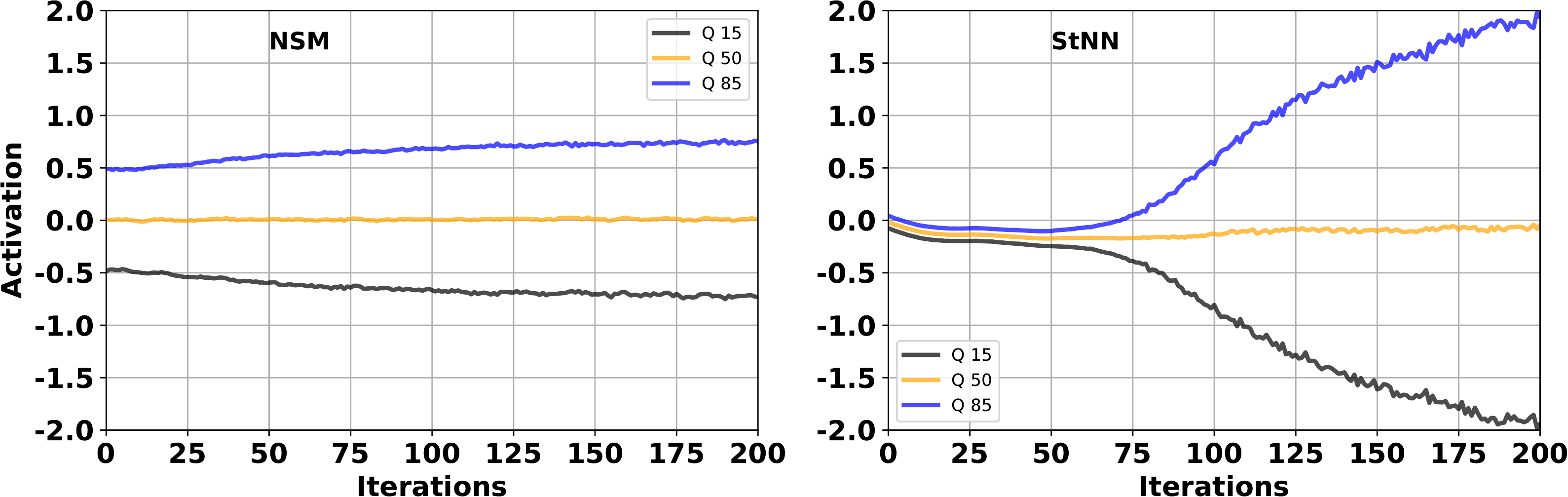}
\caption{NSM mitigates internal covariate shift. $15$th, $50$th and
$85$th percentiles of the input distribution to the last hidden layer (similarly
to Fig. $1$ in~\cite{Ioffe_Szegedy15}). The internal covariate shift is 
visible in the {StNN} as the input distributions change significantly during the
learning. The self normalizing effect in NSM performs 
weight normalization, which is known to mitigate this shift and speed up learning.
Each iteration corresponds to one mini-batch update (100 data samples per mini-batch,
$20000$ data samples total).}%
\label{fig:percentiles}
\end{figure}

The results of this experiment are shown in Table~\ref{tab:pi_mnist_errors}. 
The $15$th, $50$th and $85$th percentiles of the input distributions to the last 
hidden layer during training is shown in Fig.~\ref{fig:percentiles}. The evolution
of the distribution in the NSM case is more stable, suggesting that NSMs indeed
prevent internal covariate shift.

Both the speed of convergence and accuracy within $200$ iterations are higher in the NSM compared to the {StNN}.
The higher performance in the NSM is achieved using inference  dynamics that are simpler than the {StNN} (sign activation function compared to a sigmoid activation function) and using binary random variables.


\begin{table}
  \centering
  \caption{Classification error on the permutation invariant
  MNIST task (test set). Error is estimated by averaging test errors over 
  $100$ samples (for NSMs) and over the $50$ last epochs.}
  \label{tab:pi_mnist_errors} 
  \begin{tabular}{l l l}  
    \toprule
    Data set & Network & NSM \\
    \midrule
    PI MNIST & NSM 784--300--300--300--10 &1.36 \%    \\ 
    PI MNIST & {StNN} 784--300--300--300--10 &1.47 \%    \\ 
    PI MNIST & NSM scaled 784--300--300--300--10 &1.38 \%   \\
    \bottomrule
  \end{tabular}\\ 
\end{table}

\subsection{Robustness to Weight Fluctuations}
The decoupling of the weight matrix as in 
$\mathbf{v}_i = \beta_i \frac{\mathbf{w}_{i}}{||\mathbf{w}_{i}||}$ introduces
several additional advantages in learning machines.
During learning, the distribution of the weights for a layer tend to remain more
stable in NSM compared to the {StNN} (SI Fig. \ref{fig:w3}).
This feature can be exploited to mitigate saturation at the boundaries of fixed
range weight representations (\emph{e.g.} in fixed-point representations or 
memristors).
Another subtle advantage from an implementation point of view is that the 
probabilities are invariant to positive scaling of the weights, \emph{i.e.} 
$\frac{\alpha \mathbf{w}_{i}}{||\alpha \mathbf{w}_{i}||} = 
\frac{\mathbf{w}_{i}}{||\mathbf{w}_{i}||}$ for $\alpha\ge 0$. 
Table~\ref{tab:pi_mnist_errors} shows that NSM with weights multiplied by a constant 
factor $.1$ (called NSM scaled in the table) during inference did not significantly affect the classification
accuracy.
This suggests that the NSM can be robust to common mode fluctuations that may affect the rows of the weight matrix. 
Note that this property does not hold for ANNs with standard activation functions
(relu, sigmoid, tanh), and the network performance is lost by such scaling (for more 
details see SI~\ref{app:rob}).

\FloatBarrier
\subsection{Supervised Classification Experiments: MNIST Variants}
We validate the effectiveness of NSMs in supervised classification experiments
on MNIST~\cite{LeCun_etal98}, EMNIST~\cite{Cohen_etal17}, N-MNIST~\cite{Orchard_etal15}, and DVS Gestures data sets (See Methods) using
convolutional architecture.
For all data sets, the inputs were converted to $-1/+1$ binary in a deterministic fashion using the function defined in equation~\eqref{eq:hopfield-activation}. For the MNIST variants we trained all
the networks for $200$ epochs presenting at each epoch the entire dataset. For testing the accuracy
of the networks we used the entire test dataset sampling each minibatch $100$ times.

NSM models with Gaussian noise (gNSM) and Bernouilli noise (bNSM) converged to similar or better accuracy compared to the architecturally equivalent deterministic models.
The results for MNIST, EMNIST and N-MNIST are given in Table~\ref{table:mnist_errors}, where we compare with the deterministic counterpart convolutional neural network (see Table~\ref{table:mnist_arch} in the SI). {In addition we compared with 
a binary (sign non-linearity) deterministic network (BD), a binary deterministic network with weight normalization (wBD), a stochastic network (noisy rectifier~\cite{Bengio_etal13})
(SN), and a deterministic binary network (BN). We trained the first three networks
using a Straight-Through Estimator~\cite{Bengio_etal13} (STE) and the latter one using $\text{erf}$ function in the backward pass only (\emph{i.e.}, gradients computed on the erf function). The architecture of all these four networks is the same as the
NSM's.
The training process was the same as for the NSM networks and the results are given in Table~\ref{table:mnist_compare}. From these results we can conclude that NSM training procedure provides better performance than the STE and normalization of binary deterministic networks trained with a similar way as NSM (\emph{e.g.}, BN).}

\begin{table}[!htpb]
  \caption{(Top) Classification error on MNIST datasets.
      Error is estimated by averaging test errors over $100$ samples (for NSMs), $5$ runs, and over the $10$ last epochs. Prefix, d-deterministic, b-Bernouilli, g-Gaussian.
      (Bottom) Comparison of networks on MNIST classification task.
      The NSM variations Bernoulli (bNSM) and Gaussian (gNSM) are compared with an NSM trained with a Straight-Through Estimator instead of the proposed training algorithm, a deterministic binary (sign non-linearity) network (BD), a BD with weight normalization enabled (wBD), a stochastic network (noisy rectifier) (SN) and a binary network (BN).
      For more details see section $3.3$ in the main text. }
      \label{table:mnist_compare} \label{table:mnist_errors}

      \begin{center}
      \begin{tabular}{l l l l}  
      \toprule
      \textbf{Dataset} & \textbf{dCNN} & \textbf{bNSM} & \textbf{gNSM} \\
      \midrule
      MNIST            & 0.880\%   & 0.775 \%& 0.805\%    \\ 
      EMNIST           & 6.938\%   & 6.185 \%& 6.256\%    \\ 
      NMNIST           & 0.927\%   & 0.689 \%& 0.701\%    \\ 
      \bottomrule
                       &           &         &            \\ 
                       &           &         &            \\ 
                       &           &         &            \\ 
                       &           &         &            \\ 
      \end{tabular} 
      \begin{tabular}{cccccccc}
      \toprule
      \textbf{Model} & bNSM    & gNSM    & bNSM (STE) & BD     & wBD    & SN     &  BN \\ 
      \midrule
      \textbf{Error} & $0.775$ & $0.805$ & $2.13$ & $3.11$ & $2.72$ & $2.05$ &  $1.10$  \\
      \bottomrule
    \end{tabular}
    \end{center}
\end{table}



\subsection{Supervised Classification Experiments: CIFAR10{/100}}
We tested the NSM on the CIFAR10 and {CIFAR100} dataset of natural images.
We used the model architecture described in~\cite{Salimans_Kingma16} and added
an extra input convolutional layer to convert RGB intensities into binary values. 
The NSM non-linearities are sign functions given by
equation~\eqref{eq:hopfield-activation}. We used the
Adam~\cite{kingma:2014adam} optimizer, with initial learning rate $0.0003$ and
we trained for $200$ epochs using a batch size of $100$ {over the entire 
CIFAR10/100 data sets ($50\mathrm{K}/10\mathrm{K}$ images for training and testing respectively). The test error was computed after each epoch and by running $100$
times each batch (MC samples) with different seeds. Thus classification was made
on the average over the MC samples.}
After $100$ epochs we started decaying the learning rate linearly and we changed the first moment from $0.9$ to $0.5$.
The results are given in Table~\ref{table:cifar10_error}.
For the NSM networks we tried two different types of initialization. First,
we initialized the weights with the values of the already trained deterministic 
network weights.
Second and in order to verify that the initialization does not affect dramatically the training, we initialized the NSM without using any pre-trained weights.
In both cases the performance of the NSM was similar as it is indicated in Table~\ref{table:cifar10_error}. 
We compared with the counterpart deterministic implementation using
the exact same parameters and same additional input convolutional layer. 

\begin{table}[!htpb]
  \caption{Classification error on CIFAR10/{CIFAR100}. {Error is estimated by sampling $100$ times each mini-batch (MC samples) and finally averaging over
  all $100$ samples (for NSMs)}, $5$ runs and over the $10$ last 
    epochs. Prefix, d-deterministic, b-blank-out, g-Gaussian. {The $^\ast$ 
    indicates a network that has not been initialized with pre-trained weights 
    (see main text).}}
  \label{table:cifar10_error}
  \centering
  \begin{tabular}{lll}
    \toprule
    \textbf{Dataset} & \textbf{Model} & \textbf{Error}\\
    \midrule
    CIFAR10{/100} & bNSM & $9.98 \%$ / $34.85 \%$\\
    CIFAR10{/100} & gNSM & $10.35 \%$ / $34.84 \%$ \\
    CIFAR10{/100} & dCNN & $10.47 \%$ / $34.37 \%$ \\
    {CIFAR10/100} & bNSM$^{\ast}$ & $9.94 \%$ / $35.19 \%$ \\
    {CIFAR10/100} & gNSM$^{\ast}$ & $9.81\%$ / $34.93 \%$ \\
    \bottomrule
  \end{tabular}
\end{table}

\subsection{Supervised Classification Experiments: DVS Gestures}

\begin{table}
    \centering
    \caption{Classification error on DVS Gestures data set.
    Error is estimated by averaging test errors over $100$ samples and over the 
    $10$ last epochs. Prefix, d-deterministic, b-blank-out, g-Gaussian.}
   \label{table:dvs_errors}
    \begin{tabular}{l l l}  
      \toprule
      \textbf{Dataset} & \textbf{Model} & \textbf{Error} \\
      \midrule                  
         DVS Gestures & IBM EEDN &8.23\%    \\ 
         DVS Gestures & bNSM &8.56\%    \\ 
         DVS Gestures & gNSM &8.83\%    \\ 
         DVS Gestures & dCNN &9.16\%    \\ 
     \bottomrule
    \end{tabular}
\end{table}

Binary neural networks, such as the NSM are particularly suitable for discrete
or binary data. 
Neuromorphic sensors such as Dynamic Vision Sensors (DVS) that output streams of events fall into this category and
can transduce visual or auditory spatiotemporal patterns into parallel, microsecond-precise
streams of events ~\cite{Liu_Delbruck10}.

Amir~\emph{et al.} recorded DVS Gesture data set using a Dynamical Vision Sensor (DVS), comprising $1342$ instances of a set of $11$ hand and arm gestures,
collected from $29$ subjects under $3$ different lighting conditions.
Unlike standard imagers, the DVS records streams of events that signal the 
temporal intensity changes at each of its $128 \times 128$ pixels.
The unique features of each gesture are embedded in the stream of events. 
To process these streams, we closely follow the pre-processing
in~\cite{Amir_etal17}, where event streams were downsized to $64\times 64$ and
binned in frames of $16\mathrm{ms}$. 
The input of the neural was formed by 6 frames (channels) and only ON (positive
polarity) events were used.
Similarly to~\cite{Amir_etal17}, $23$ subjects are used for the training set, and the remaining $6$ subjects are reserved for testing.
We note that the network used in this work is much smaller than the one used in~\cite{Amir_etal17}.

We adapted a model based on the all convolutional networks of~\cite{Springenberg_etal14}.
Compared to the original model, our adaptation includes an additional group of
three convolutions and one pooling layer to account for the larger image size
compared to the CIFAR10 data set used in~\cite{Springenberg_etal14} and a 
number of output classes that matches those of the DVS Gestures data set 
(11 classes). 
See SI Tab.~\ref{tab:allcnn} for a detailed listing of the layers. We trained
the network for $200$ epochs using a batch size $100$. For the NSM network we
initialized the weights using the converged weights of the deterministic
network. This makes learning more robust and causes a faster convergence.  

We find that the smaller models of~\cite{Springenberg_etal14} (in terms of 
layers and number of neurons) are faster to train and perform equally well when
executed on GPU compared to the EEDN used in~\cite{Amir_etal17}. 
The models reported in Amir~\emph{et al.} were optimized for implementation in
digital neuromorphic hardware, which strongly constrains weights, connectivity
and neural activation functions in favor of energetic efficiency.

\section{Conclusions}

Stochasticity is a powerful mechanism for improving the computational features of 
neural networks, including regularization and Monte Carlo sampling.
This work builds on the regularization effect of stochasticity in neural 
networks, and demonstrates that it naturally induces a normalizing effect on 
the activation function.
Normalization is a powerful feature used in most modern deep neural networks
\cite{Ioffe_Szegedy15,Ren_etal16,Salimans_Kingma16}, and mitigates internal
covariate shift.
Interestingly, this normalization effect may provide an alternative mechanism 
for divisive normalization in biological neural networks~\cite{Carandini_Heeger11}.

Our results demonstrate that NSMs can (i) outperform standard stochastic networks
on standard machine learning benchmarks on convergence speed and accuracy,
and (ii) perform close to deterministic feed-forward networks when data is of 
discrete nature.
This is achieved using strictly simpler inference dynamics, that are well suited
for emerging nanodevices, and argue strongly in favor of \emph{exploiting} 
stochasticity in the devices for deep learning.
Several implementation advantages accrue from this approach: it is an online
alternative to batch normalization and dropout, it mitigates saturation at the
boundaries of fixed range weight representations, and it confers robustness 
against certain spurious fluctuations affecting the rows of the weight matrix. 

Although feed-forward passes in networks can be implemented free of multiplications,
the weight update rule is more involved as it requires multiplications, 
calculating the row-wise $L2$-norms of the weight matrices, and the derivatives
of the $\mathrm{erf}$ function. 
However, these terms are shared for all connections fanning into a neuron, 
such that the overhead in computing them is reasonably small.
Furthermore, based on existing work, we speculate that approximating the 
learning rule either by hand~\cite{Neftci_etal17} or automatically~\cite{Andrychowicz_etal16}
can lead to near-optimal learning performances while being implemented with
simple primitives.




\newpage
\bibliographystyle{plain}
\bibliography{biblio}

\newpage
\section*{Supplementary Information}

\subsection{{Table of Abbreviations}}

\begin{table}[!htpb]
  \caption{Abbreviations used in the main text and in the SI.}
  \label{table:cifar10_error}
  \centering
  \begin{tabular}{ll}
    \toprule
    \textbf{Abbreviation} & \textbf{Definition} \\
    \midrule
    NSM & Neural Sampling Machine \\
    b/gNSM & Bernoulli/Gaussian NSM \\
    S2M & Synaptic Sampling Machine \\
    {StNN} & Stochastic Neural Network \\
    BNN & Binary Neural Networks \\
    DVS & Dynamic Vision Sensor \\
    BD & Deterministic Binary Network with $\sgn$ as non-linearity \\
    wBD & Same as BD with weight normalization enabled \\
    SN & Stochastic network with noisy rectifier \\
    BN & Binary network\\
    STE & Straight-Through Estimator \\
    \bottomrule
  \end{tabular}
\end{table}

\subsection{{Computation of Gradients in NSM Computational Graph}}
\label{app:graph}

\begin{figure}[!htpb]
    \centering
    \includegraphics[width=0.3\textwidth]{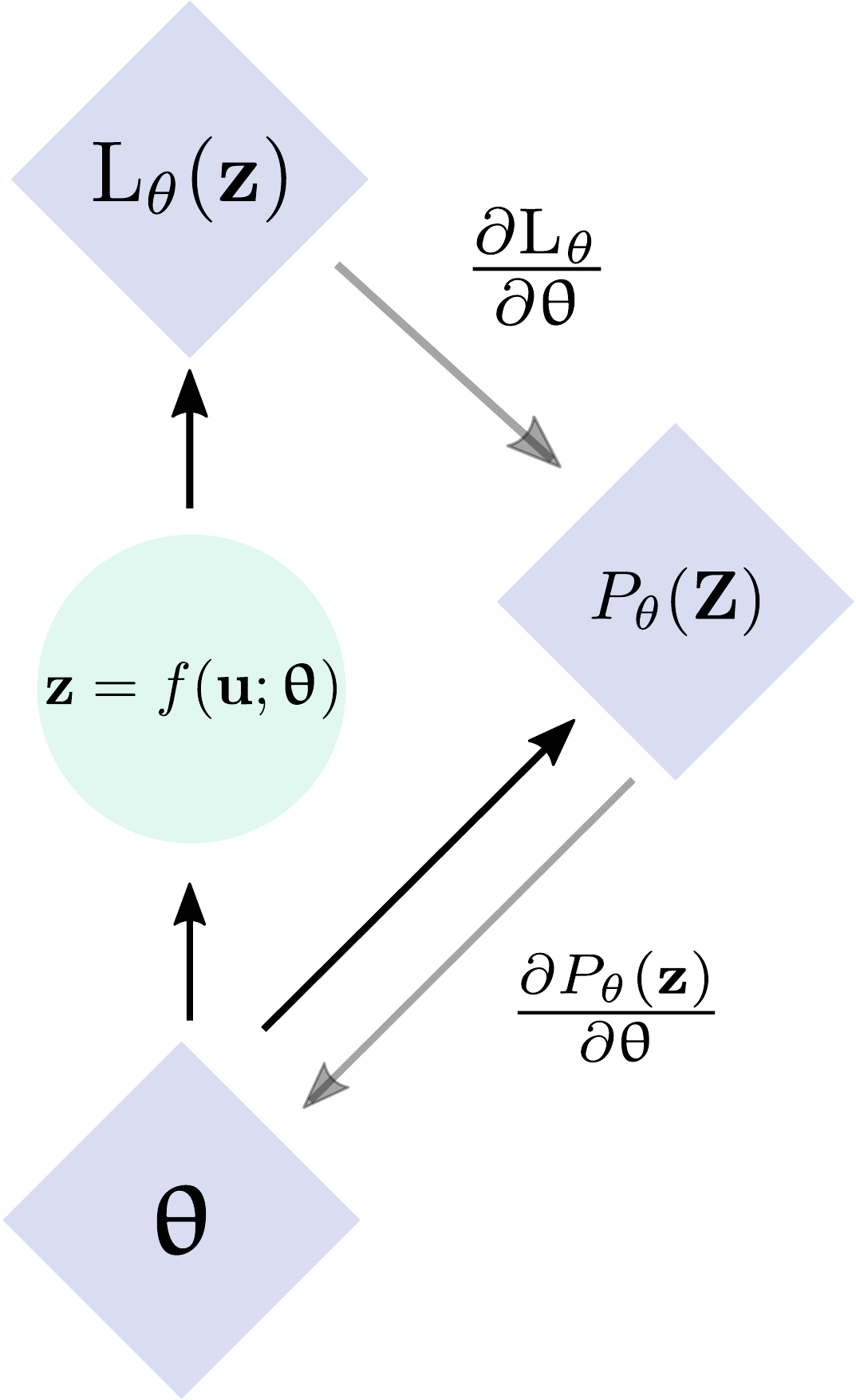}
    \caption{{Gradient estimation in NSM computation graph. For the NSM 
    network the gradient $\nabla_{\theta}L(x)$ is computed via back-propagation 
    on the probability $P_{\theta}({\bf z})$ only in the backward pass (see equation~\eqref{eq:general_activation_function} and main text). 
    The light-green node indicates
    a stochastic discrete node that propagates the activity of units to the next
    layer only in the forward pass. The parameters here are $\boldsymbol{\theta} = ({\bf w},  \boldsymbol{\beta})$ (see main text).}}
    \label{fig:comp_graph}
\end{figure}

\subsection{NSM in Convolutional Neural Networks}
CNN perform state-of-the-art in several visual, auditory and natural language tasks by assuming prior 
structure to the connectivity and the weight matrices~\cite{LeCun_etal98,Goodfellow_etal16}.
The NSM with stochastic neurons can be similarly extended to the convolution operation as follows
(bias parameters omitted):
\begin{equation}
  \begin{split}
    u_{ijk} &= Conv(\mathbf{w}_k,\mathbf{z})\\
            &= \sum_{q=1}^{Q}\sum_{m=1}^{H}\sum_{n=1}^{V} (\xi_{i+m,j+n,q} + a_{ijk}) w_{m,n,q,k} z_{i+m,j+n,q} 
  \end{split}
\end{equation}
where $Q$ is the number of input channels and $H,V$ are height and width of the filter, respectively.
In the case of neural stochasticity, existing software libraries of the convolution can be used.
In contrast, synaptic stochasticity, requires modification of such libraries due to the sharing of the
filter parameters. While it is possible to do so, we have not observed significant differences in using
neural or synaptic stochasticity. Therefore only neural stochasticity is used for convolution operations.
Similarly to the case without convolutions, the activation probability becomes:
\begin{equation}
    \begin{split}
      P(z_{ijk} = 1| \mathbf{z}) & \frac12 \left(1 + \mathrm{erf}\left( Conv(\mathbf{v}_k,\mathbf{z})  \right)\right), \\
      \text{with } \mathbf{v}_i &= \beta_{ijk}    \frac{\mathbf{w}_{k}}{||\mathbf{w}_{k}||}, 
    \end{split}
\end{equation}
where,
\begin{equation}
  ||\mathbf{w}_{k}|| = \sqrt{\sum_{m=1}^{H}\sum_{n=1}^{V} \sum_{q=1}^{Q} w_{m,n,q,k}^2}.
\end{equation}

\subsection{{Derivation of Gradients (Equations~\eqref{eq:derivative_L1} and~\eqref{eq:derivative_L2})}}
In this section we derive equations~\eqref{eq:derivative_L1} and~\eqref{eq:derivative_L2}. Therefore, if we differentiate through 
${\bf v}_i = \beta_i \frac{{\bf w}_i}{||{\bf w}_i||}$, we obtain equation~\eqref{eq:derivative_L1} from 
\begin{align}
    \label{eq:derivation_l1}
    \frac{\partial \mathcal{L}}{\partial \boldsymbol{\beta}_i} &= \frac{\partial \mathcal{L}}{\partial {\bf v}_i} \frac{\partial {\bf v}_i}{\partial \boldsymbol{\beta}_i} \nonumber \\
    &= \frac{\partial \mathcal{L}}{\partial {\bf v}_i}\frac{{\bf w}_i}{||{\bf w}_i||} \nonumber \\
    &=  \frac{\sum_{j}w_{ij} \partial_{v_{ij}} \mathcal{L}}{||{\bf w}_i||}.
\end{align}
And it is obvious that equation~\eqref{eq:derivative_L1} is equation~\eqref{eq:derivation_l1}. For obtaining equation~\eqref{eq:derivative_L2}
we have
\begin{align}
    \label{eq:derivation_l2}
    \frac{\partial \mathcal{L}}{\partial {\bf w}_i} &= \frac{\partial \mathcal{L}}{\partial {\bf v}_i} \frac{\partial {\bf v}_i}{\partial {\bf w}_i} \nonumber \\
    &= \frac{\partial \mathcal{L}}{\partial {\bf v}_i}\frac{\beta_i \partial \frac{{\bf w}_i}{||{\bf w}_i||}}{\partial {\bf w}_i} \nonumber \\
    &= \frac{\partial \mathcal{L}}{\partial {\bf v}_i} \bigg(\beta_i \frac{\frac{\partial {\bf w}_i} {\partial {\bf w}_i} ||{\bf w}_i|| - {\bf w}_i \frac{\partial ||{\bf w}_i||}{\partial {\bf w}_i}  }{||{\bf w}_i||^2} \bigg)  \nonumber \\ 
    &= \frac{\partial \mathcal{L}}{\partial {\bf v}_i} \bigg(\beta_i \frac{||{\bf w}_i|| - {\bf w}_i \frac{\sum_{j}^{}w_{ij}}{||{\bf w}_i||}  }{||{\bf w}_i||^2} \bigg) \nonumber \\
    &=  \frac{\partial \mathcal{L}}{\partial {\bf v}_i} \frac{\beta_i ||{\bf w}_i||}{||{\bf w}_i||^2}
    - \frac{\partial \mathcal{L}}{\partial {\bf v}_i} \frac{\beta_i {\bf w}_i \sum_{j}^{}w_{ij}}{||{\bf w}_i||^2 ||{\bf w}_i||} \nonumber  \\
    &=  \frac{\partial \mathcal{L}}{\partial {\bf v}_i} \frac{\beta_i}{||{\bf w}_i||} -
    \frac{\beta_i}{||{\bf w}_i||^2} \frac{\partial \mathcal{L}}{\partial {\bf v}_i} \frac{{\bf w}_i}{||{\bf w}_i||}  \sum_{j}^{} w_{ij}  \nonumber \\
    &= \frac{\beta_i}{||{\bf w}_i||} \partial_{v_{ij}} \mathcal{L} - \frac{\beta_i {\bf w}_i }{||{\bf w}_i||^2} 
    \frac{\sum_{j}^{} w_{ij} \partial_{v_{ij}} \mathcal{L}}{||{\bf w}_i||} \nonumber \\
       &= \frac{\beta_i}{||{\bf w}_i||} \partial_{v_{ij}} \mathcal{L} - \frac{\beta_i}{||{\bf w}_i||^2} 
        {\bf w}_i \partial_{\beta_i} \mathcal{L}. 
\end{align}
Therefore, we have derived equation~\eqref{eq:derivative_L2} (which is equation~\eqref{eq:derivation_l2}).

\subsection{{Robustness to Weight Fluctuations}}
\label{app:rob}

\begin{figure}[!htpb]
    \centering
    \includegraphics[width=0.8\textwidth]{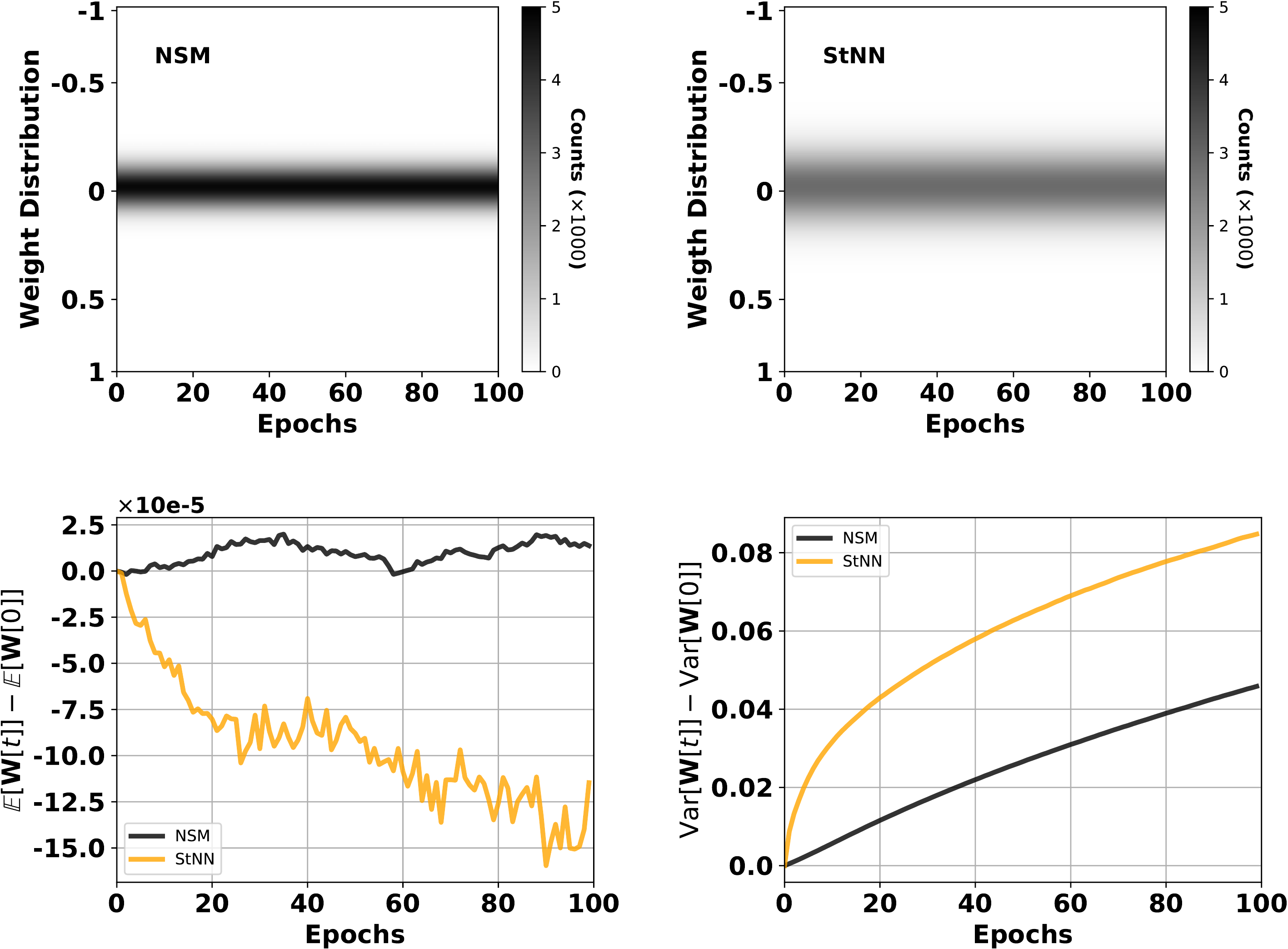}
\caption{Evolution of $W_3$ ({\emph{i.e.}, weights of the third layer)} weight distributions during learning, normalized to initial values (top row). In the NSM, the scale of the weights is controlled by the factors $\beta_i$. This renders the weights during learning more stable (left panel, top row) compared to the sigmoid neural network (right panel, top row), which tends to grow at a faster rate.
The mean of NSM remains close to zero (black line, bottom left panel) in comparison to the mean 
of the StNN (yellow line). Similar to the mean, the variance of NSM (black line, bottom right panel) grows slower and remains smaller than that of StNN (yellow line, bottom right panel).}
\label{fig:w3}
\end{figure}

{The decoupling of the weight matrix (\emph{i.e.}, $\mathbf{v}_i = \beta_i \frac{\mathbf{w}_{i}}{||\mathbf{w}_{i}||}$) introduces a robustness to weights
fluctuation. During learning, the distribution of the weights for each layer 
tends to remain more stable in NSM compared to StNN. See for instance
Figure~\ref{fig:w3}, where in the top row the evolution of weights distribution
of the third layer (${\bf W}_3$) is shown for the NSM and the StNN, respectively. 
It is apparent that the distribution of NSM weights is more narrow and remains
concentrated around its mean (low variance). 
On the other hand, the variance of the weight distribution in larger in the StNN.
The same results are  illustrated in the two bottom panels where the mean of the weights of the third
layer over training is subtracted from the mean of the initial weights. 
We observe that the NSM is more robust and the mean remains almost steady (left
panel) in comparison to StNN. The same phenomenon is observed also in the case of 
standard deviation (right panel), where the NSM's standard deviation increases 
slightly in comparison to StNNs}.

\subsection{{Training NSMs with BinConcrete}}
\label{app:con}

This section details how the NSM can be trained using the
BinConcrete distribution instead of propagating gradients through the activation probability function (see main text and SI~\ref{app:graph}).
In the forward pass, the probability is computed using equation~\eqref{eq:general_activation_function}
and then passed to the BinConcrete~\cite{Maddison_etal16} given by 
the following equation
\begin{align}
    \label{eq:concrete}
    X &= \sigma \Big(\frac{L + \log(\alpha)}{\lambda}\Big),
\end{align}
where $\alpha$ is the probability we have already computed, $\sigma$ is the sigmoid 
function, $L$ is the Logistic distribution ($\log(U) - \log(1 - U)$, where $U$ is 
the uniform distribution in the $[0, 1]$) and $\lambda$ is the temperature term. 
In our experiments, we assume that $\lambda = 1$.
In the backward pass, the gradients are computed through equation~\eqref{eq:concrete} instead of equation~\eqref{eq:general_activation_function}.

\newpage

\subsection{{N}-MNIST}
The N-MNIST data set uses the same digits as contained in MNIST \cite{Orchard_etal15}.
The digits were presented to an event-based camera  that detects temporal contrast (ATIS), and their output was recorded.
The data set consists of binary files, each containing the information of a single digit. Each file contains four arrays of equal length describing: the
$x$ coordinate and $y$ coordinates of an event, the polarity (on or off) and 
the timestamp of the event. For this network, only the positive polarity events
were extracted. Ten $34\times 34$ frames of zeros were created for each digit
and the maximum timestamp was divided by $10$ to obtain the frame length. 
For each digit, an entry in the frame corresponding to the $x$ and $y$ 
coordinates of the events extracted inside the designated frame time was 
changed from $0$ to $1$.  This was repeated for each of the ten frames. 
Test error results were obtained averaging test errors across the last $5$ 
epochs and over $X$ separate runs with different seed values.

\subsection{Simulation Details}
\label{app:sim_details}
The source code for this work is written in Python and Pytorch~\cite{paszke:2017} 
and it is available online under the GPL license at \url{https://github.com/nmi-lab/neural_sampling_machines}.
We ran all the experiments on two machines:
\begin{enumerate}
    \item A Ryzen ThreadRipper with $64 \mathrm{GB}$ physical memory running Arch Linux,
    Python $3.7.4$, Pytorch $1.2.0$ and GCC $9.1.0$, equipped with three Nvidia GeForce GTX $1080$ Ti
    GPUs. 
    \item A Intel i$7$ with $64 \mathrm{GB}$ physical memory running Arch Linux, Python $3.7.3$, 
    Pytorch $1.0.1$, and GCC $8.2.1$, equipped with two Nvidia GeForce RTX $2080$ Ti GPUs.
\end{enumerate}

\subsection{MNIST, EMNIST, NMNIST Neural Networks}
\label{app:mnist_net}

\begin{table}[!htpb]
  \caption{Convolutional neural network used for MNIST, EMNIST, NMNIST data 
  sets.}
  \label{table:mnist_arch}
  \centering
  \begin{tabular}{lll}
    \toprule
    Layer Type & \# Channels & $x$, $y$ dimension  \\
    \midrule
    Raw Input & $1$ & $28$ \\
    $5 \times 5$ Conv & $32$ & $24$ \\
    $2 \times 2$ Max Pooling (stride 2) & $32$ & $12$ \\
    $5 \times 5$ Conv & $64$ & $8$ \\
    $2 \times 2$ Max Pooling (stride 2) & $64$ & $4$ \\
    $1024 \times 512$ FC & $1024$ & $1$ \\
    Softmax output      & $10$  & $1$\\
    \bottomrule
  \end{tabular}
\end{table}

\newpage

\subsection{DVS Gestures Neural Network}

\begin{table}[!htpb]
  \caption{ All convolutional neural network used for the DVS Gestures dataset.}
  \label{tab:allcnn}
  \centering
  \begin{tabular}{lll}
    \toprule
    Layer Type & \# Channels \& Dimensions  \\
    \midrule
    Input (ON events) & 6 & $64\times 64$ \\
    $3 \times 3$ Conv & 96 & $64\times 64$ \\
    $3 \times 3$ Conv & 96 & $64\times 64$ \\
    $3 \times 3$ Conv & 96 & $64\times 64$ \\
    $2 \times 2$ Max Pooling (stride 2) & 96 & $32\times 32$ \\
    $3 \times 3$ Conv & 192 & $32\times 32$ \\
    $3 \times 3$ Conv & 192 & $32\times 32$ \\
    $3 \times 3$ Conv & 192 & $32\times 32$ \\
    $2 \times 2$ Max Pooling (stride 2) & 192 & $16\times 16$ \\
    $3 \times 3$ Conv & 256 & $16\times 16$ \\
    $3 \times 3$ Conv & 256 & $16\times 16$ \\
    $3 \times 3$ Conv & 256 & $16\times 16$ \\
    $2 \times 2$ Max Pooling (stride 2) & 256 & $8\times 8$ \\
    $3 \times 3$ Conv & 256 & $8\times 8$ \\
    $1 \times 1$ Conv & 256 & $8\times 8$ \\
    $1 \times 1$ Conv & 256 & $8\times 8$ \\
    Global average pool & 256 & $1\times 1$\\
    Softmax       & 11  & $1\times 1$\\
    \bottomrule
  \end{tabular}
\end{table}

\newpage

\subsection{Weights Statistics for MNIST Classification}\label{sec:concrete}

\begin{figure}[!htpb]
    \centering
    \includegraphics[width=0.8\textwidth]{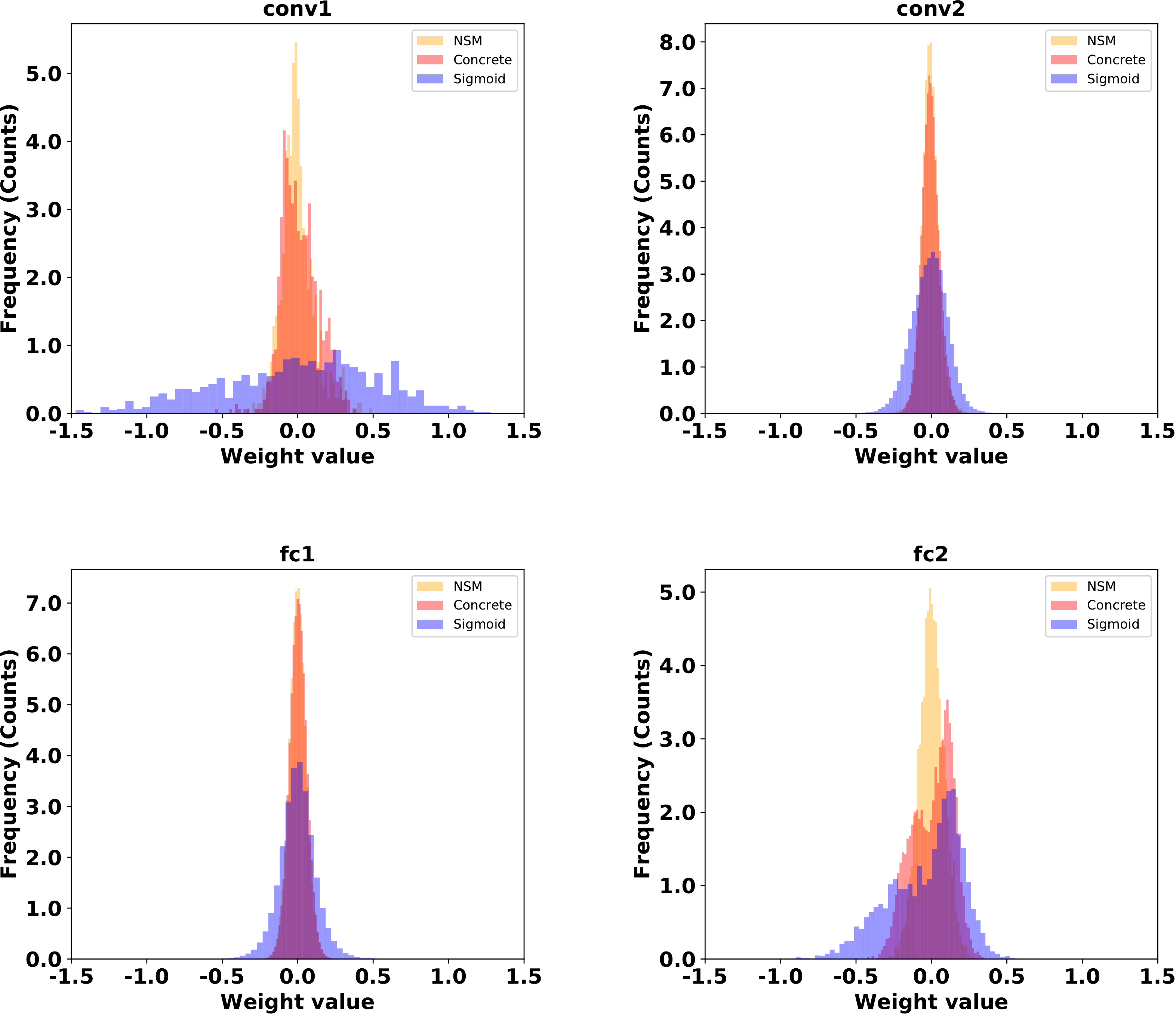}
    \caption{Histograms of Weights on MNIST Classification. The weights of four layers,
    convolutional layers 1 and 2 and fully connected layers 1 and 2 (see SI~\ref{app:mnist_net})
    for three different neural networks, NSM (yellow), NSM trained with Concrete Distribution (red, see
    SI~\ref{app:con}), 
    and Deterministic one with sigmoid as non-lineariry. }
    \label{fig:hists}
\end{figure}

\begin{figure}[!htpb]
    \centering
    \includegraphics[width=0.8\textwidth]{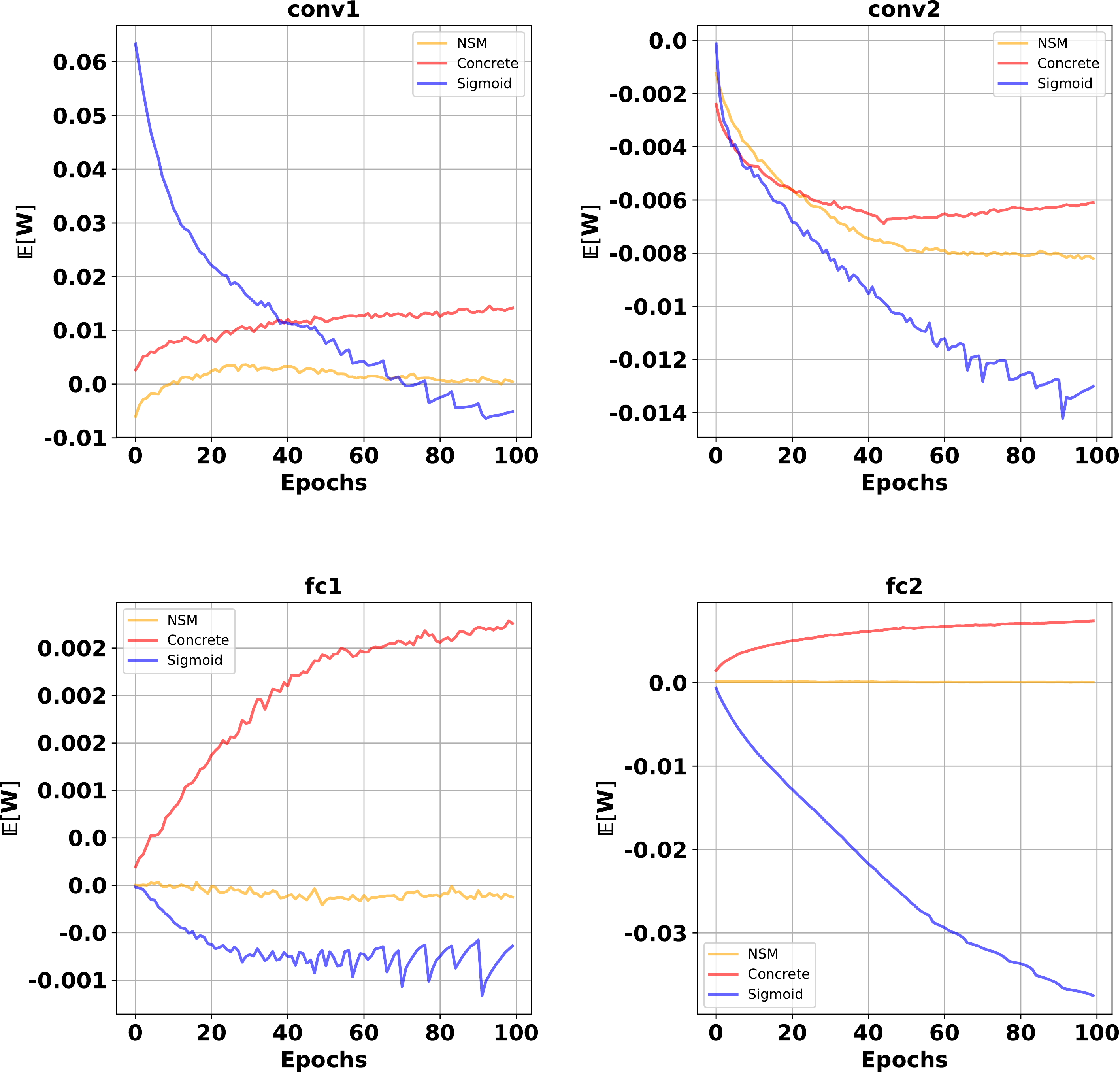}
     \caption{Mean of Weights on MNIST Classification. Mean weights of the four layers of 
     the neural network used in main text for MNIST classification. Convolutional layers 1 and 2
     and fully connected layers 1 and 2 (see SI~\ref{app:mnist_net}). 
     Weights of three different neural networks are presented here, NSM (yellow), NSM trained 
     with BinConcrete Distribution (red, see SI~\ref{app:con}), and a Deterministic one
     with sigmoid as non-linearity (blue). }
    \label{fig:mean_w}
\end{figure}

\begin{figure}[!htpb]
    \centering
    \includegraphics[width=0.8\textwidth]{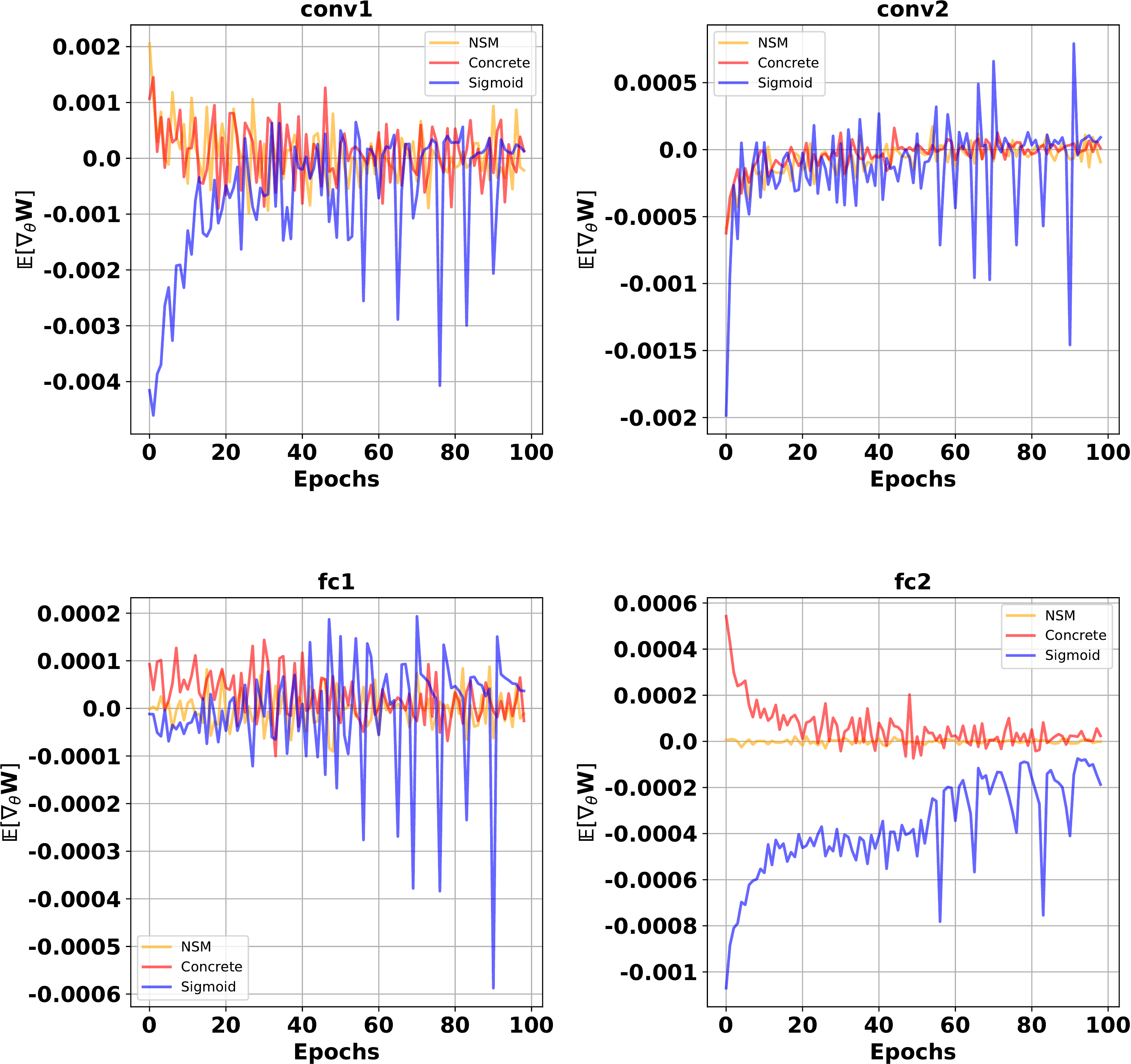}
     \caption{Mean of Weights Gradients on MNIST Classification. Mean of weights gradients of
     the four layers of the neural network used in main text for MNIST classification. Convolutional 
     layers 1 and 2
     and fully connected layers 1 and 2 (see SI~\ref{app:mnist_net}). 
     Weights of three different neural networks are presented here, NSM (yellow), NSM trained 
     with BinConcrete Distribution (red, see SI~\ref{app:con}), and a Deterministic one
     with sigmoid as non-linearity (blue). }
    \label{fig:mean_g}
\end{figure}

\begin{figure}[!htpb]
    \centering
    \includegraphics[width=0.8\textwidth]{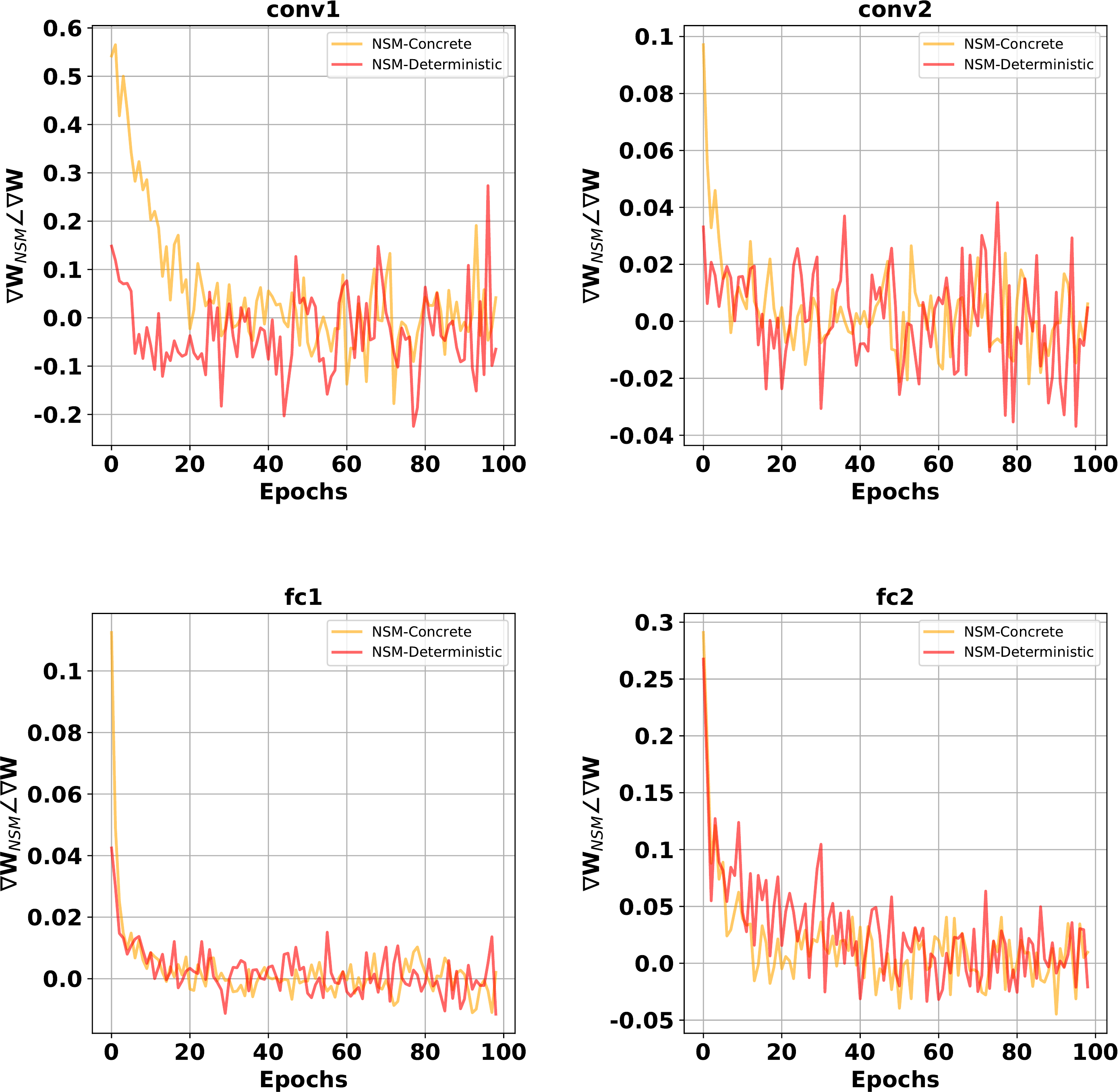}
     \caption{Angles (cosine similarity) of Gradients on MNIST Classification. The angles of gradients of weights 
     of three networks are compared with each other. Four layers, convolutional layers 1 and 2 and
     fully connected layers 1 and 2 (see SI~\ref{app:mnist_net}) are shown in this figure. 
     Two cases are illustrated in this figure, (i) NSM against NSM trained with BinConcrete (yellow, 
     see SI~\ref{app:con}), (ii) NSM against a Deterministic network with sigmoid as non-linearity
     (red).}
    \label{fig:angles_g}
\end{figure}

In this section we provide some statistics on the weights of the convolutional neural network
used in the MNIST classification task. The network architecture is given in SI~\ref{app:mnist_net} 
and the results of the classification task are given in Table~\ref{table:mnist_errors} in main 
text. First, we provide the histogram of the weights after training on the MNIST data set for three
different types of networks. 
An NSM network, an NSM trained using the BinConcrete distribution and 
a deterministic network with sigmoid function as non-linearity. For more details about the networks 
see the main text, and SI~\ref{app:mnist_net} and~\ref{app:con}. 
Histograms are illustrated in Figure~\ref{fig:hists}. Then we measured the expected value of the weights
for each layer and for each network as well as the mean gradients of the weights. Those results are 
shown in Figures~\ref{fig:mean_w} and~\ref{fig:mean_g}, respectively. Finally, we show the angles between 
the gradients of NSM weights and NSM trained with BinConcrete (yellow) and NSM and Deterministic (orange)
in Figure~\ref{fig:angles_g}. These results indicate that the NSM and the NSM with BinConcrete express 
similar behavior during training. On the other hand, the deterministic network has larger weights and 
develops larger gradient steps (blue lines in Figures~\ref{fig:mean_w} and~\ref{fig:mean_g}).

\newpage

\end{document}